# Classification of the Chess Endgame problem using Logistic Regression, Decision Trees, and Neural Networks

Mahmoud S. Fayed

King Saud University

Riyadh, Saudi Arabia

msfayed@ksu.edu.sa

## Abstract

In this study we worked on the classification of the Chess Endgame problem using different algorithms like logistic regression, decision trees and neural networks. Our experiments indicates that the Neural Networks provides the best accuracy (85%) then the decision trees (79%). We did these experiments using Microsoft Azure Machine Learning as a case-study on using Visual Programming in classification. Our experiments demonstrates that this tool is powerful and save a lot of time, also it could be improved with more features that increase the usability and reduce the learning curve. We also developed an application for dataset visualization using a new programming language called Ring, our experiments demonstrates that this language have simple design like Python while integrates RAD tools like Visual Basic which is good for GUI development in the open-source world.

## Introduction

In a puzzle game like a Chess, the game could have the next three stages [1-3].

1. Chess Opening

2. Chess Middlegame

3. Chess Endgame

The Chess Endgame is a state where we have a few numbers of pieces left on the board [4-5]. There are many studies that uses different algorithms to predict the result of the Chess endgame [6-8]. In this study we will apply a group of algorithms on the chess endgame problem (King Rook vs King) and test the accuracy of the multiclass classifiers [9-10].

The remainder of this paper is organized as follows. Section 2. Describe the related works. Section 3. Illustrates the problem definition. Section 4. Demonstrates using Microsoft Azure Machine Learning to implement the different models. while Section 5. Demonstrates using the Ring programming language to build the Chess Endgame Application which provides Data Set Visualization and a use interface for game result prediction. Finally, we present the future work and the conclusion in Section 6.

## Related Work

The UCI Machine Learning Repository contains the King-Rook vs King data set [11]. In 1994, The Data Set authors (Bain et all) report classification results with an accuracy of 16% over the entire data set [12-13]. Their study is done using inductive logic programming [14-16]. In 2006, Lassabe et al. studied using Inductive logic programming (ILP) and Genetic Programming (GP) to create computer programs that play the chess end game, but their study does not present the classification results [17].

In 2012, Wayne Iba (Westmont College), used genetic algorithms and report accuracy between 20% and 30% using 1000 samples of the Data Set in the training stage [18]. Instead of using Inductive logic programming or Genetic Programming, the Machine Learning field contains a lot of algorithms and methods that we can try to use like Logistic Regression [19-20], Modern decision trees algorithms like Decision Jungle [21-22], Decision Forest [23-25].

Also, we can use the Artificial Neural Networks [26-32].

The next table presents the results in the previous studies

| Index | Author | Algorithm | Accuracy | Year | Ref. |
|---|---|---|---|---|---|
| 1 | Wayne Iba | Genetic Algorithms | 30% | 2012 | 18 |
| 2 | Lassabe | IPL & GP | Not Available | 2006 | 17 |
| 3 | Bain et all | Inductive Logic Programming | 16% | 1994 | 15 |

Table 1.

Visual programming languages (VPLs) are tools that can be used to create computer programs and software without writing textual code directly. In VPLs programming are done using more than one dimension, and a mix of text, shapes and icons could be used. Also some visual programming languages add the time dimension to programming [33-34].

There are VPLs in many domains and Microsoft Azure Machine Learning tool uses Visual Programming to reduce the learning curve and increase the productivity of producing machine learning models that could be intenerated to many applications [35-37]. We can use Microsoft Azure Machine learning to try many algorithms quickly and get results in short time. Also we can use a modern textual programming language like the Ring programming language to develop the application. Ring is a new language influenced with many programming language. It’s dynamic and simple like Python but, comes with Rapid Application Development tools like Visual Basic [38]. The language is designed for developing the second generation of the Programming Without Coding Technology software [39-41]. PWCT is a popular visual programming language which is used in many applications and systems including the development of the Supernova language and the Critical Nodes application for the LASCNN algorithm [42-43]. Ring is influenced by many programming languages including Lua, Python, Ruby, C, C#, Basic, QML, xBase, and Supernova. [44]

**The Chess Game**

In the Chess Game each player goal is delivering a checkmate by trapping the opponent´s king. "The possible number of chess games is so huge that no one will invest the effort to calculate the exact number.", Jonathan Schaeffer (University of Alberta). The chess board contains eight files (a,b,c,d,e,f,g,h) and eight ranks (1,2,3,4,5,6,7,8). The position of each piece on the board could be determined using the file and the rank. A common chess endgame situation is having (King-Rook) vs (King).

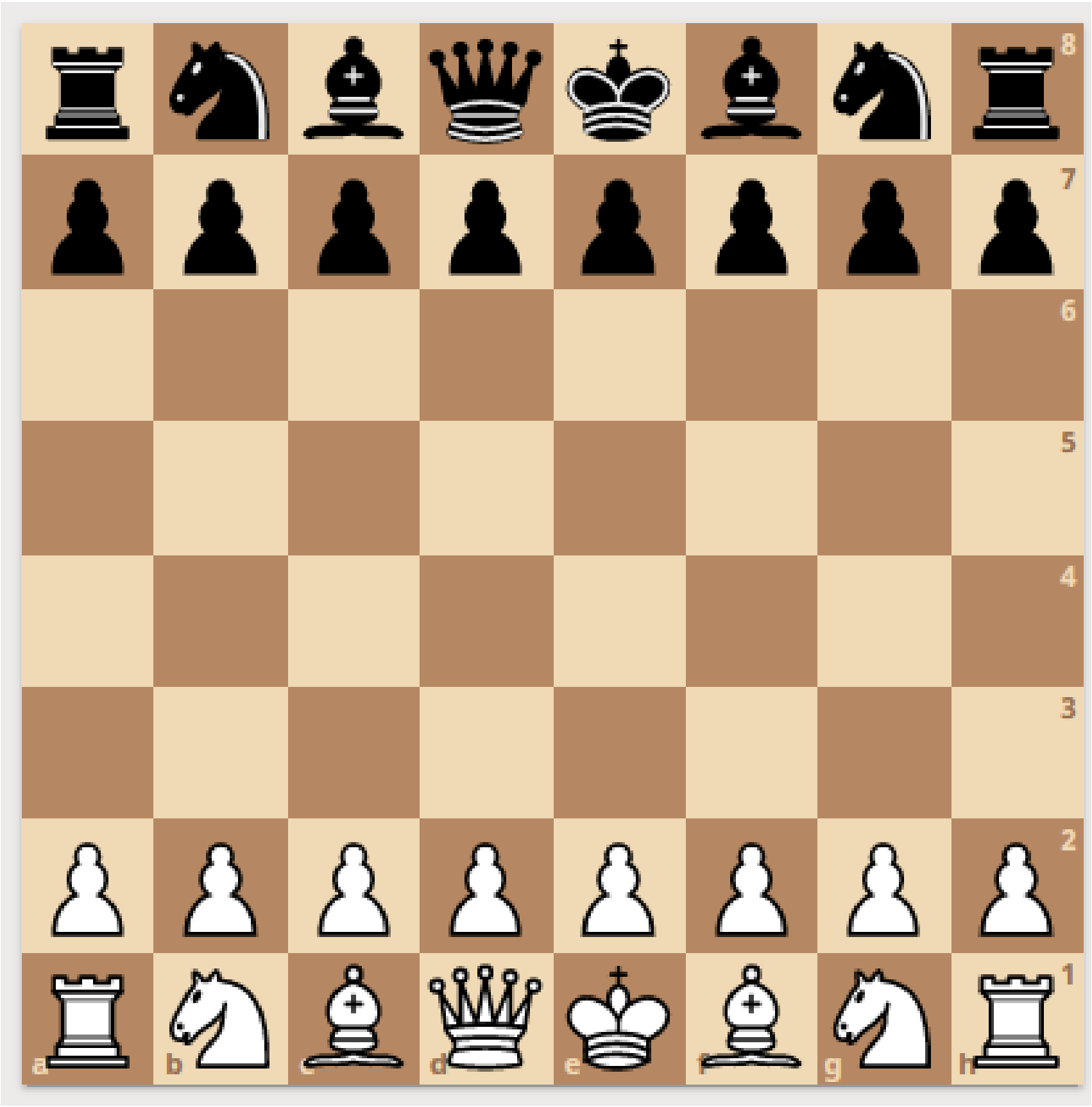

Fig. 1

## Chess (King-Rook vs King) Data Set

The data set contains 28056 instances, the date donated is 1994-06-01 and the number of attributes is six.

**Attribute Information:**

- 1. White King file (column)
  2. White King rank (row)
  3. White Rook file
  4. White Rook rank
  5. Black King file
  6. Black King rank
  7. optimal depth-of-win for White in 0 to 16 moves, otherwise drawn {draw, zero, one, two, ..., sixteen}.

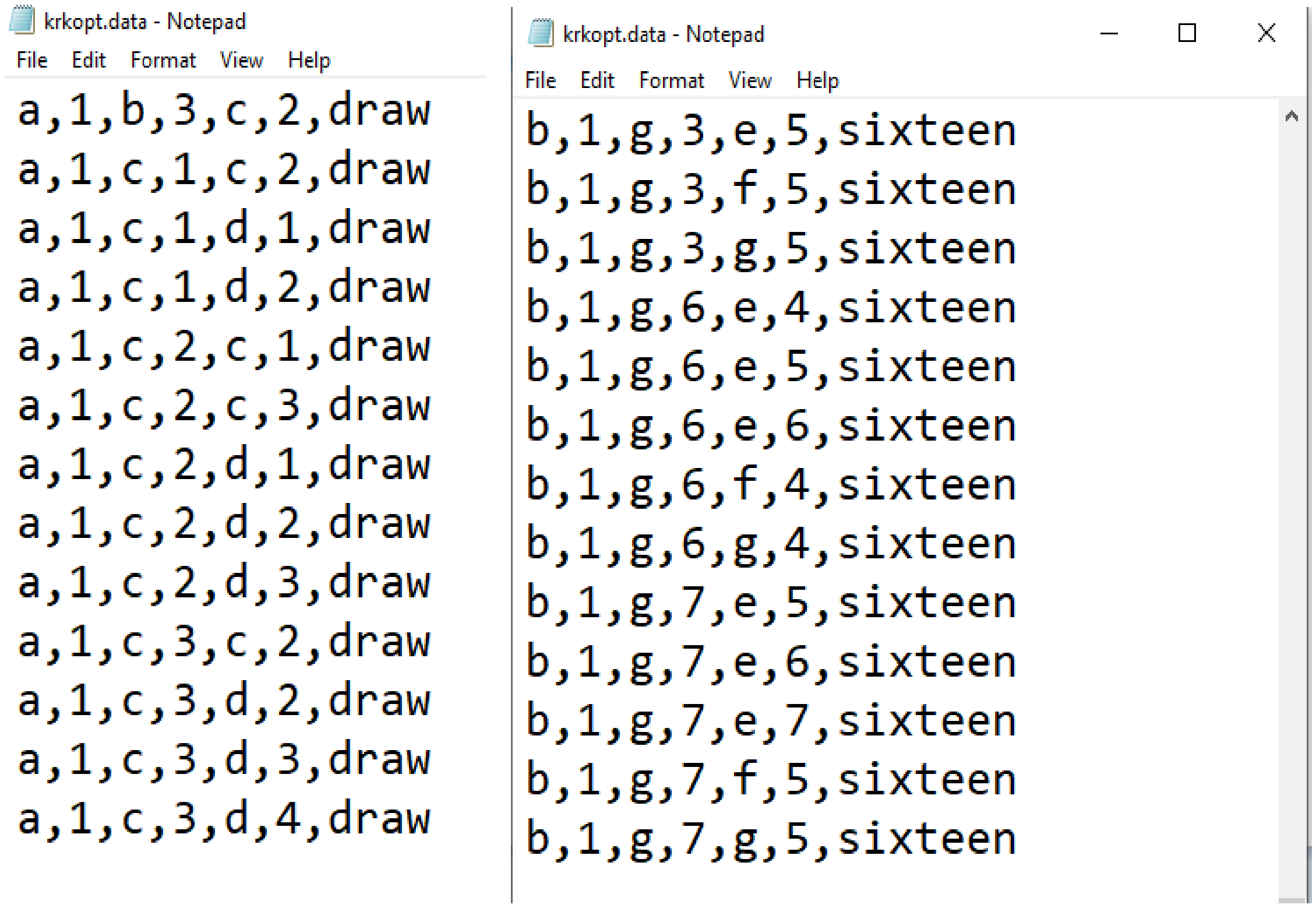
krkopt.data - Notepad
File Edit Format View Help

```
a,1,b,3,c,2,draw
a,1,c,1,c,2,draw
a,1,c,1,d,1,draw
a,1,c,1,d,2,draw
a,1,c,2,c,1,draw
a,1,c,2,c,3,draw
a,1,c,2,d,1,draw
a,1,c,2,d,2,draw
a,1,c,2,d,3,draw
a,1,c,3,c,2,draw
a,1,c,3,d,2,draw
a,1,c,3,d,3,draw
a,1,c,3,d,4,draw
```

krkopt.data - Notepad
File Edit Format View Help

```
b,1,g,3,e,5,sixteen
b,1,g,3,f,5,sixteen
b,1,g,3,g,5,sixteen
b,1,g,6,e,4,sixteen
b,1,g,6,e,5,sixteen
b,1,g,6,e,6,sixteen
b,1,g,6,f,4,sixteen
b,1,g,6,g,4,sixteen
b,1,g,7,e,5,sixteen
b,1,g,7,e,6,sixteen
b,1,g,7,e,7,sixteen
b,1,g,7,f,5,sixteen
b,1,g,7,g,5,sixteen
```

Fig. 2

The King can move one square in any direction (horizontally, vertically, or diagonally) while the Rook can move any number of squares in vertical or horizontal direction. According to the Data Set the player who will play the next move will control the Black King. In the next game the optimal result is (Draw) because the Black King can move from (C2) to (B3) and take the Rook which means (King vs. King) which is a draw.

| Piece | File | Rank |
|---|---|---|
| White King | a | 1 |
| White Rook | b | 3 |
| Black King | c | 2 |

Table. 2

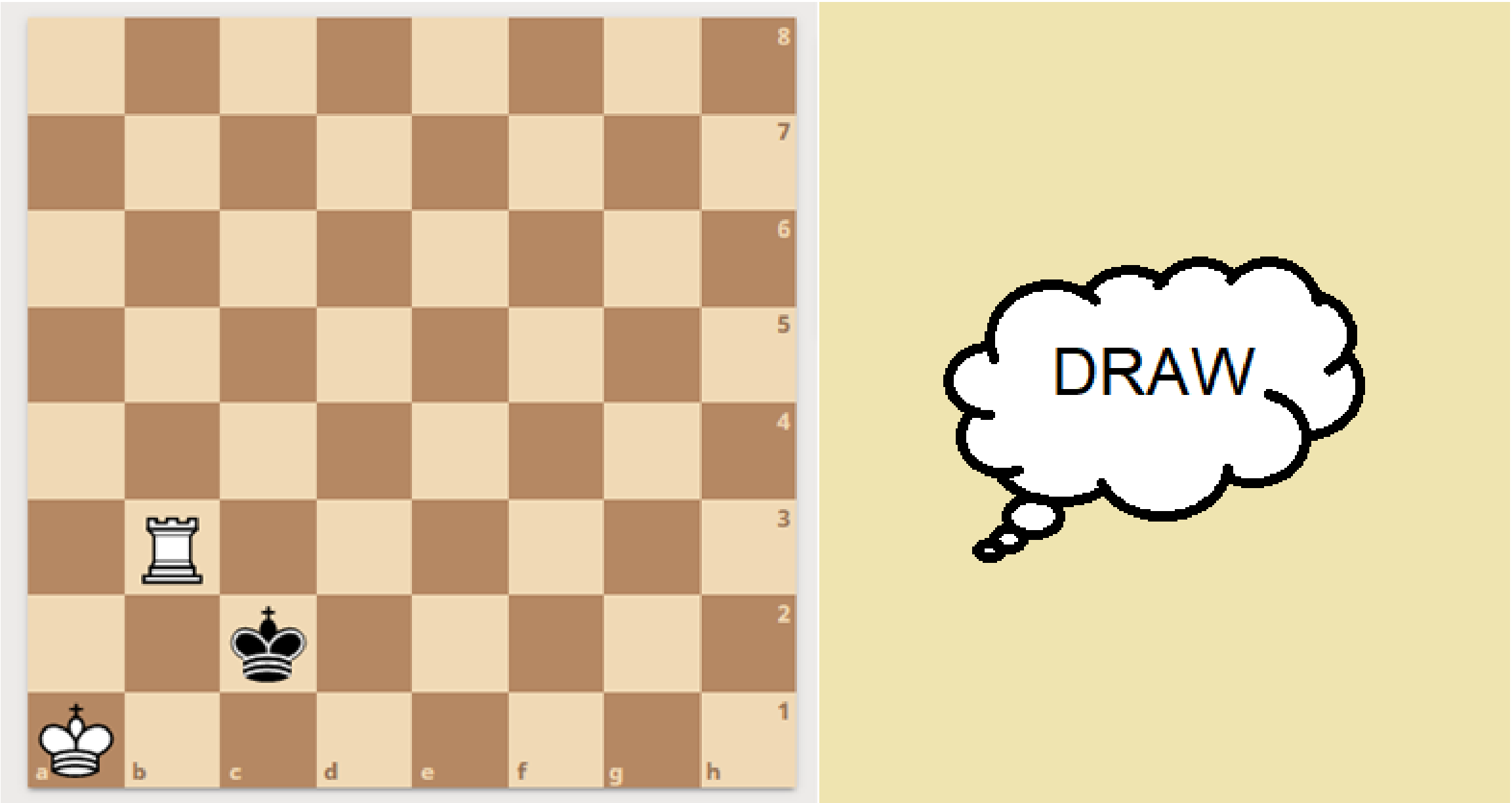

Fig. 3

In the next game, the White will win in one move

| Piece | File | Rank |
|---|---|---|
| White King | c | 1 |
| White Rook | c | 3 |
| Black King | a | 2 |

Table. 3

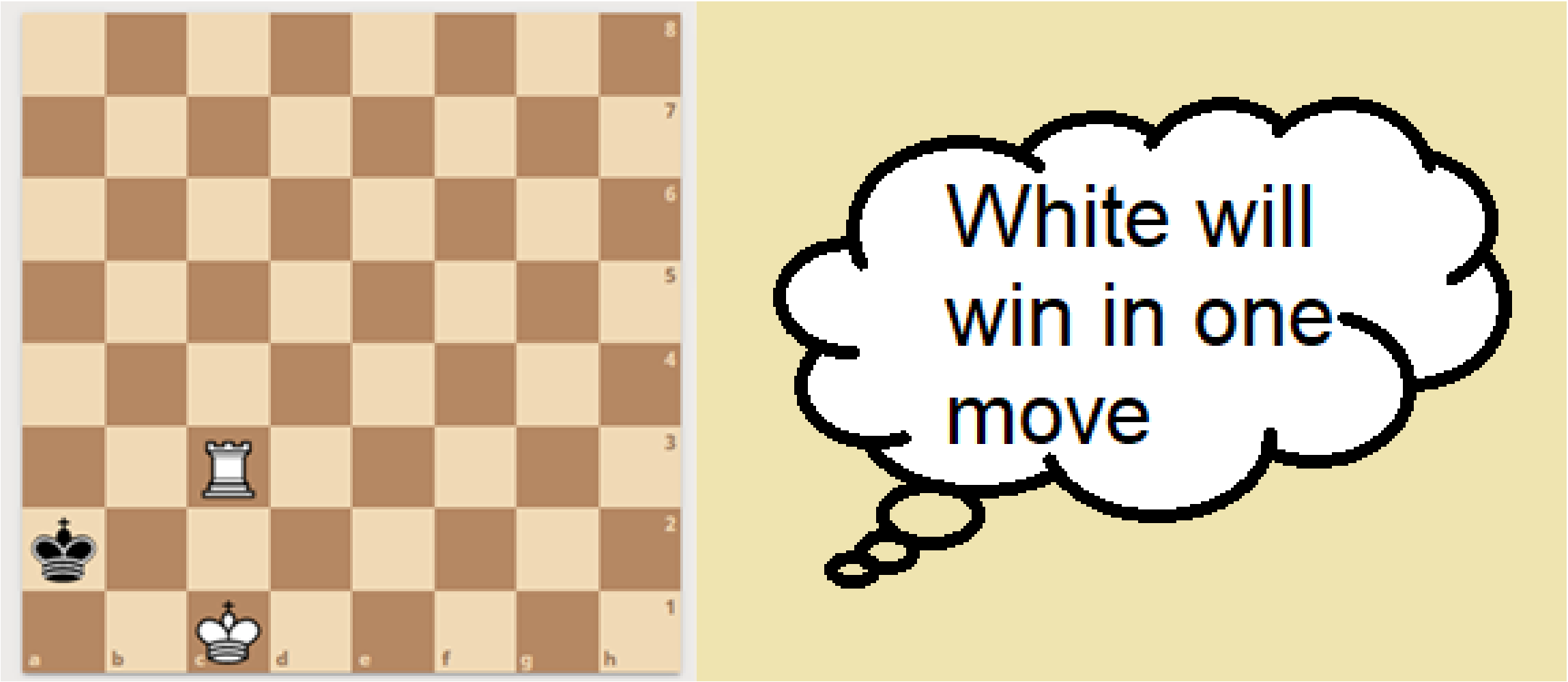

Fig. 4

The Black king must play, He can’t move to A3, B3 because of the White Rook, and can’t move to B1,B2 because of the White King. He can move only to A1

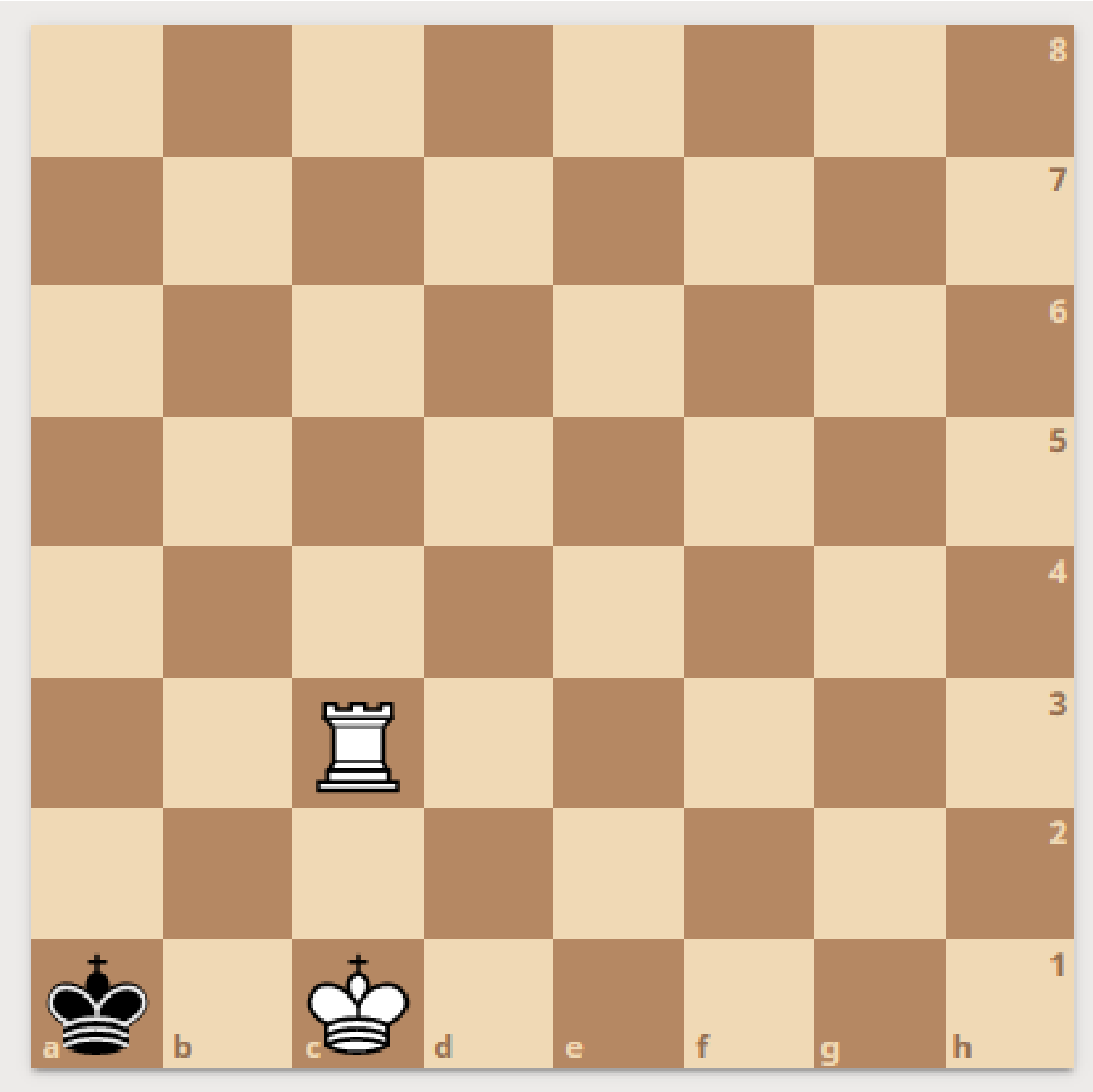

Fig. 5

In this situation, the White Rook moves from C3 to A3 and the white player wins the game (Checkmate for the Black King)

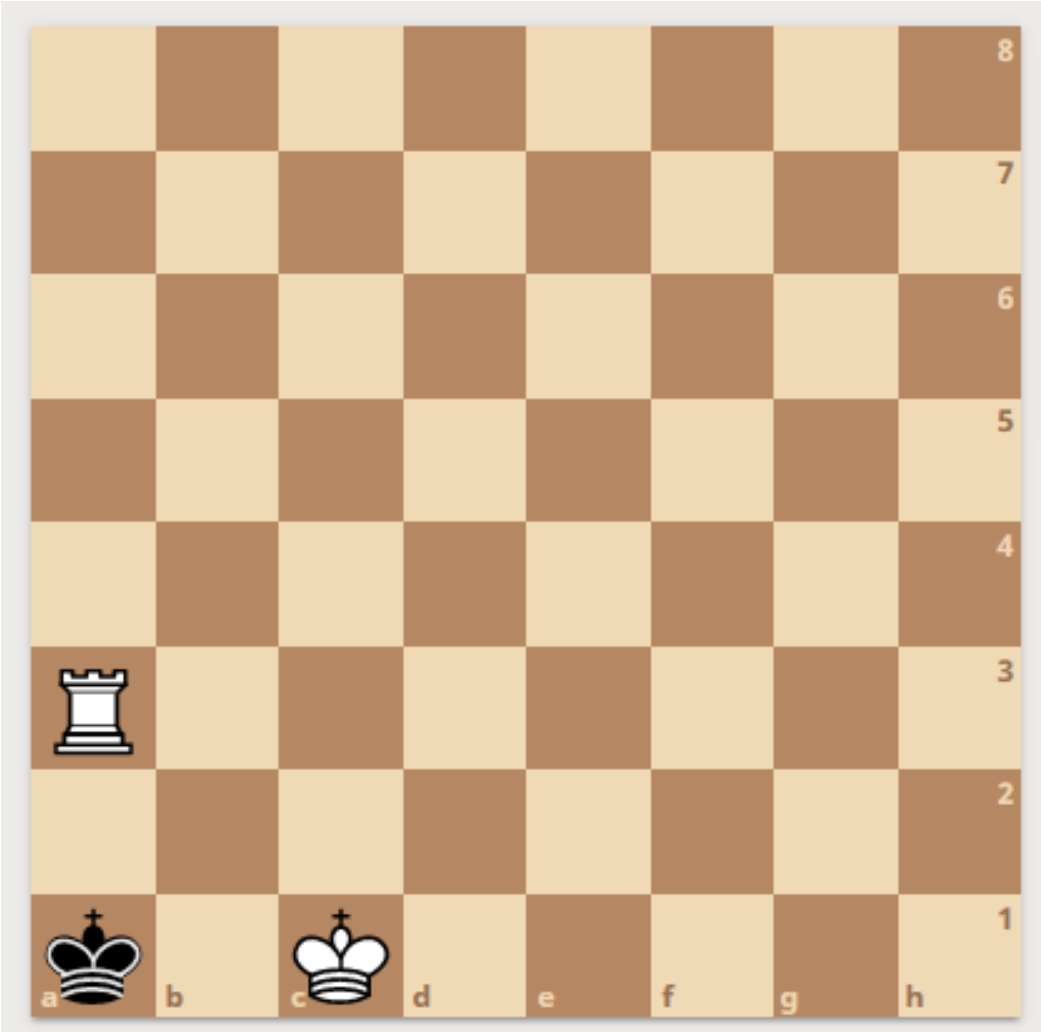

Fig. 6

**Algorithms and development tools**

This study uses the next algorithms for classification.

- Multiclass Decision Forest
- Multiclass Decision Jungle
- Multiclass Logistic Regression
- Multiclass Neural Network

The next tools are used for development.

- Microsoft Azure Machine Learning
- The Ring programming language

Steps

- Prepare and Analysis the Data Set
- Normalize the Data.
- Split the Data (Training Data & Test Data)
- Select the Model (Algorithm)
- Training
- Testing (Calculate the Accuracy)
- Compare the Results Between the different Algorithms.

## Using Azure Machine Learning

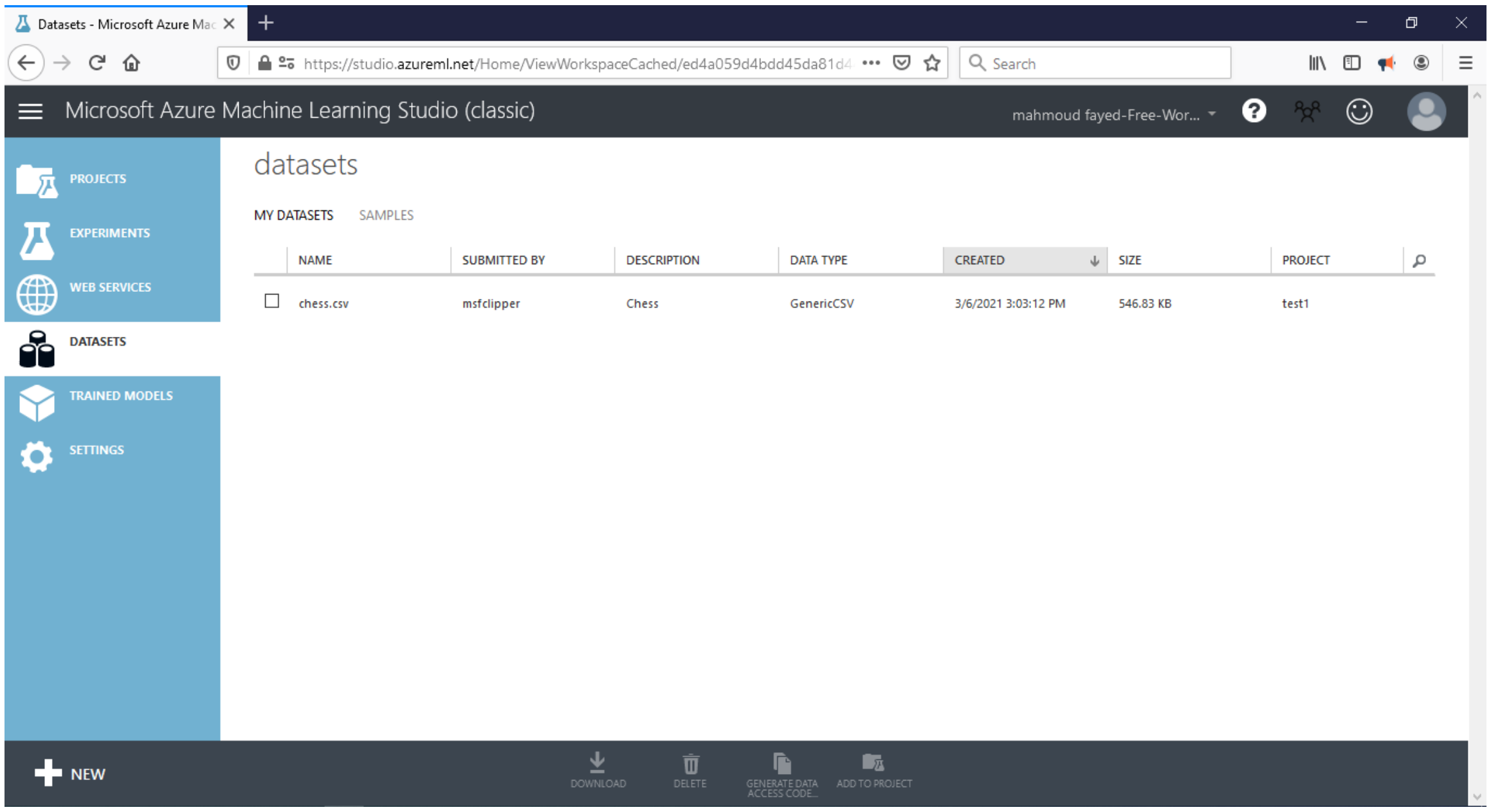

Fig. 7

## Logistic Regression

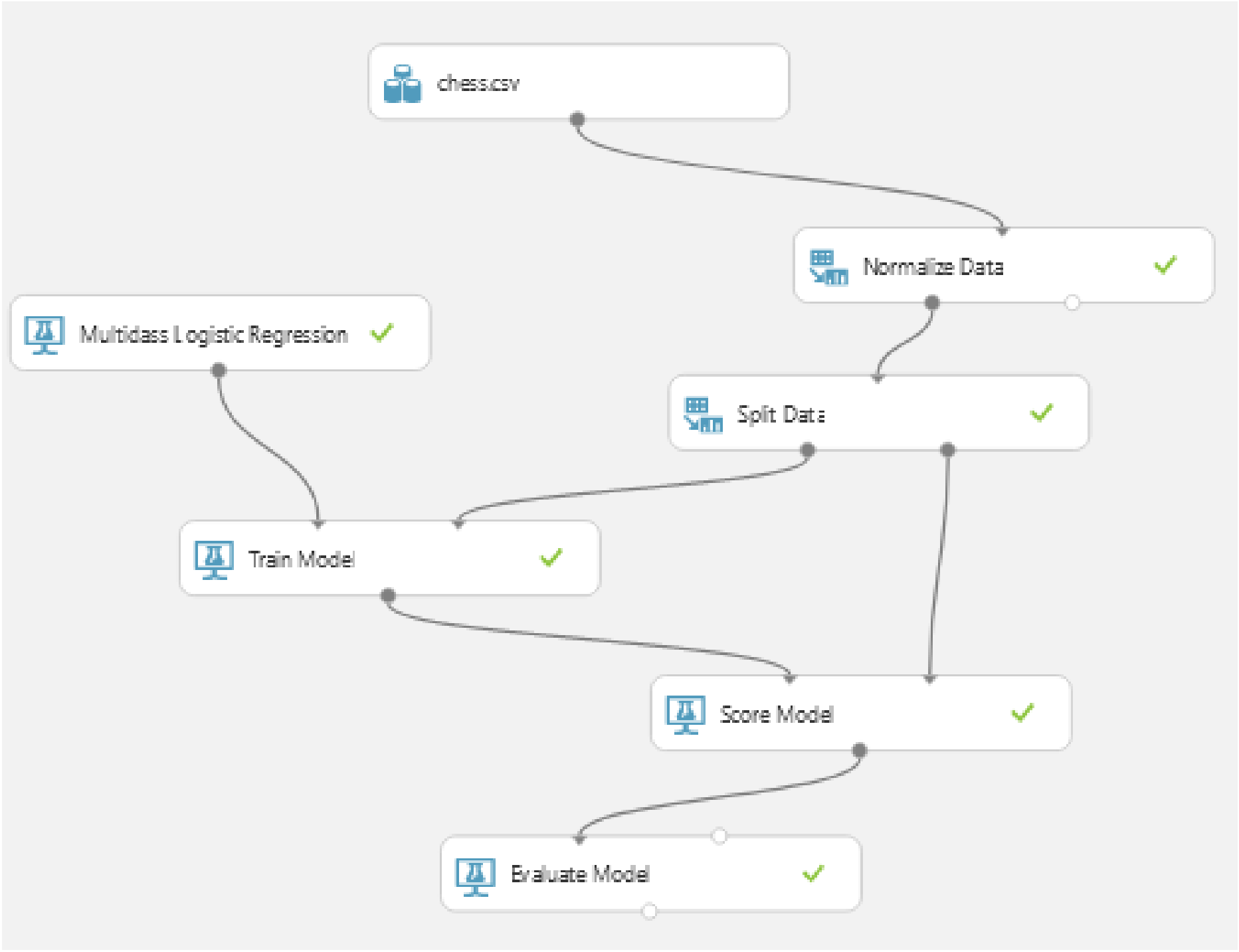

Fig. 8

Exp - LR - 3/6/2021 › Evaluate Model › Evaluation results

## Metrics

| | |
|---|---|
| Overall accuracy | 0.321255 |
| Average accuracy | 0.924584 |
| Micro-averaged precision | 0.321255 |
| Macro-averaged precision | NaN |
| Micro-averaged recall | 0.321255 |
| Macro-averaged recall | 0.282853 |

Fig. 9

**Multiclass Decision Jungle**

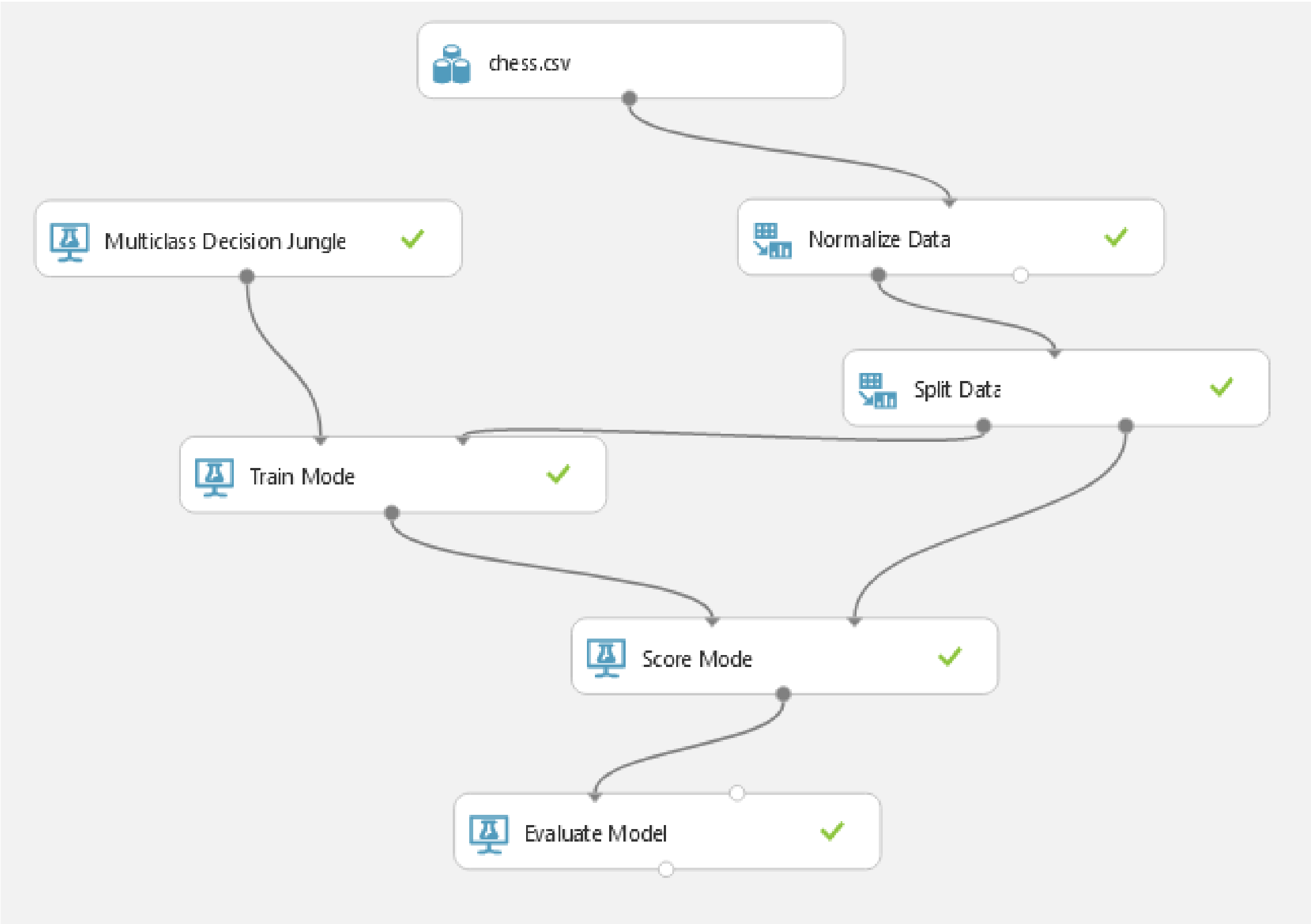

Fig. 10

Exp - DJ - 3/6/2021 › Evaluate Model › Evaluation results

◢ Metrics

| | |
|---|---|
| Overall accuracy | 0.496376 |
| Average accuracy | 0.944042 |
| Micro-averaged precision | 0.496376 |
| Macro-averaged precision | NaN |
| Micro-averaged recall | 0.496376 |
| Macro-averaged recall | 0.455527 |

Fig. 11

**Multiclass Decision Forest**

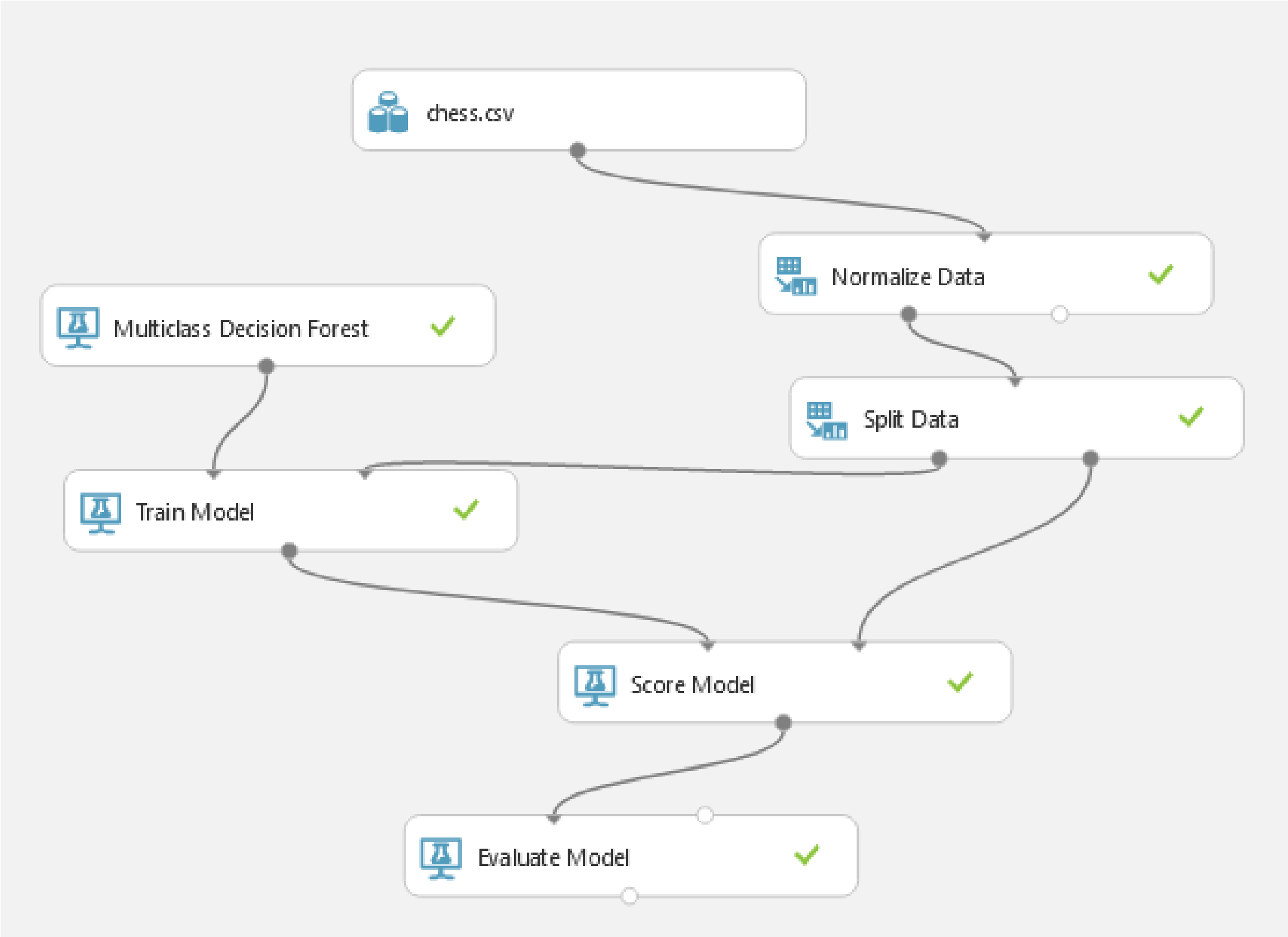

Fig. 12

Exp - DF - 3/6/2021 › Evaluate Model › Evaluation results

◢ Metrics

| | |
|---|---|
| Overall accuracy | 0.793038 |
| Average accuracy | 0.977004 |
| Micro-averaged precision | 0.793038 |
| Macro-averaged precision | 0.798891 |
| Micro-averaged recall | 0.793038 |
| Macro-averaged recall | 0.792915 |

Fig. 13

**Neural Network**

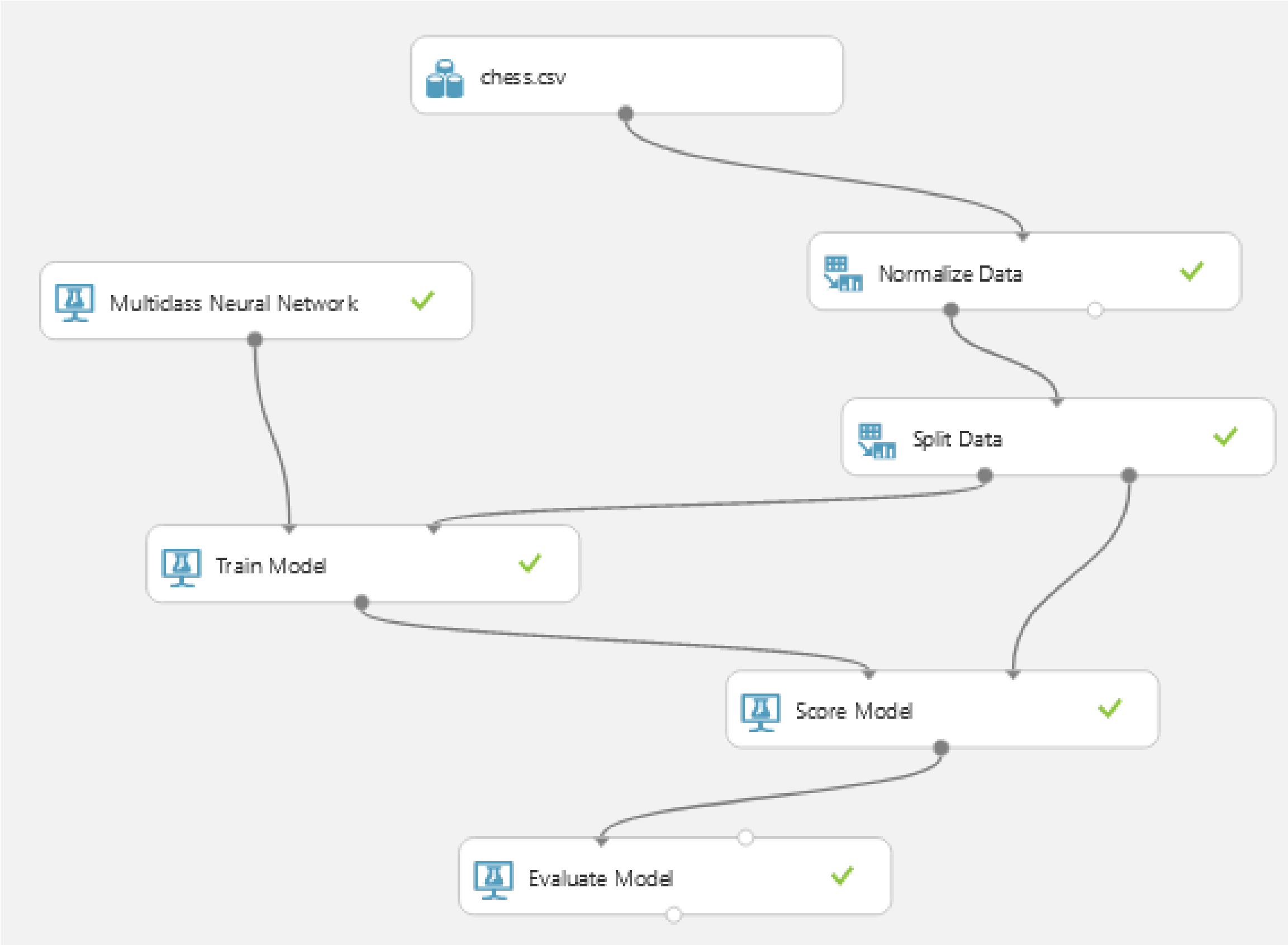

Fig. 14

Exp - NN - 3/6/2021 › Evaluate Model › Evaluation results

Metrics

| | |
|---|---|
| Overall accuracy | 0.622668 |
| Average accuracy | 0.958074 |
| Micro-averaged precision | 0.622668 |
| Macro-averaged precision | 0.631862 |
| Micro-averaged recall | 0.622668 |
| Macro-averaged recall | 0.565878 |

Fig. 15

**Early Results**

| Algorithm | Overall Accuracy | Average Accuracy |
|---|---|---|
| Multiclass Decsion Jungle | 0.496376 | 0.944042 |
| Multiclass Decision Forest | 0.793038 | 0.977004 |
| Logistic Regression | 0.321255 | 0.924584 |
| Neural Network | 0.622668 | 0.958074 |

Fig. 16

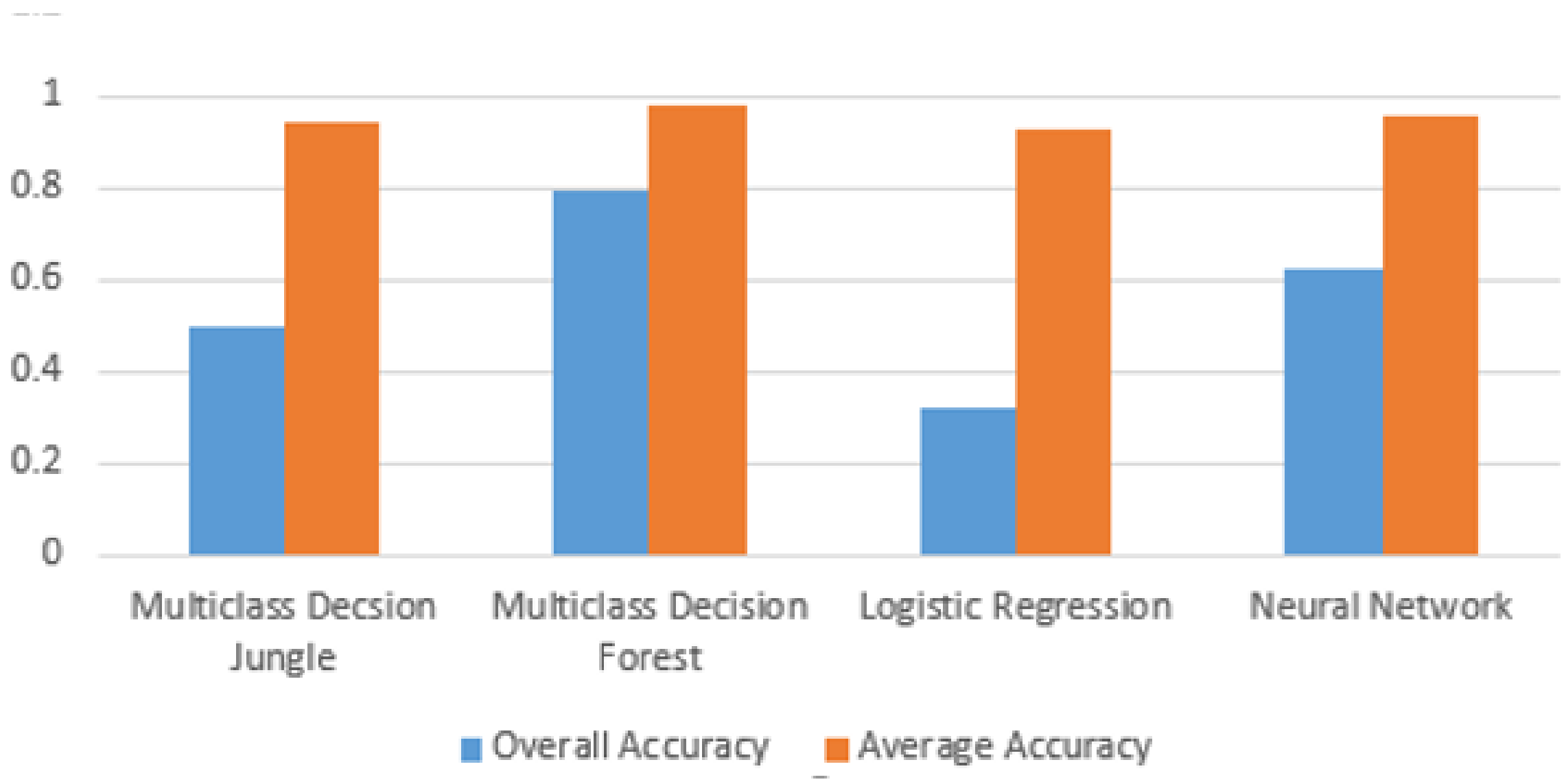

Fig. 17

## Changing the Neural Network Parameters

The Azure Machine Learning tool provide some parameters that we can set for our neural network, like the number of nodes, training time and the number of iterations. We tried many variations of these parameters and checked the change in the accuracy.

| No. of Nodes | Learning rate | No. of iterations | Overall Accuracy | Average Accuracy |
|---|---|---|---|---|
| 100 | 0.1 | 100 | 0.62 | 0.95 |
| 100 | 0.1 | 1000 | 0.66 | 0.96 |
| 100 | 0.1 | 10000 | 0.64 | 0.96 |
| 1000 | 0.1 | 100 | 0.66 | 0.96 |
| 1000 | 0.1 | 1000 | 0.71 | 0.96 |
| 1000 | 0.1 | 10000 | 0.71 | 0.96 |
| 10000 | 0.1 | 100 | 0.65 | 0.96 |
| 10000 | 0.1 | 1000 | 0.72 | 0.96 |
| 10000 | 0.1 | 10000 | 0.72 | 0.96 |
| 100 | 0.01 | 100 | 0.60 | 0.95 |
| 100 | 0.01 | 1000 | 0.67 | 0.96 |
| 100 | 0.01 | 10000 | 0.69 | 0.96 |
| 1000 | 0.01 | 100 | 0.60 | 0.95 |
| 1000 | 0.01 | 1000 | 0.71 | 0.96 |
| 1000 | 0.01 | 10000 | 0.72 | 0.96 |
| 10000 | 0.01 | 100 | 0.55 | 0.95 |
| 10000 | 0.01 | 1000 | 0.71 | 0.96 |
| 10000 | 0.01 | 10000 | 0.73 | 0.97 |
| 100 | 0.001 | 100 | 0.33 | 0.92 |
| 100 | 0.001 | 1000 | 0.60 | 0.95 |
| 100 | 0.001 | 10000 | 0.68 | 0.96 |
| 1000 | 0.001 | 100 | 0.28 | 0.92 |
| 1000 | 0.001 | 1000 | 0.58 | 0.95 |
| 1000 | 0.001 | 10000 | 0.73 | 0.97 |
| 10000 | 0.001 | 100 | 0.18 | 0.90 |
| 10000 | 0.001 | 1000 | 0.54 | 0.94 |
| 10000 | 0.001 | 10000 | 0.72 | 0.96 |

Table. 4

From these experiments, we can reach 73% accuracy using 1000 nodes, 0.001 learning rate and 10000 iterations. The next screen shot from Microsoft Azure Machine Learning demonstrates the selection of the Multiclass neural network from many other networks that we tried to test.

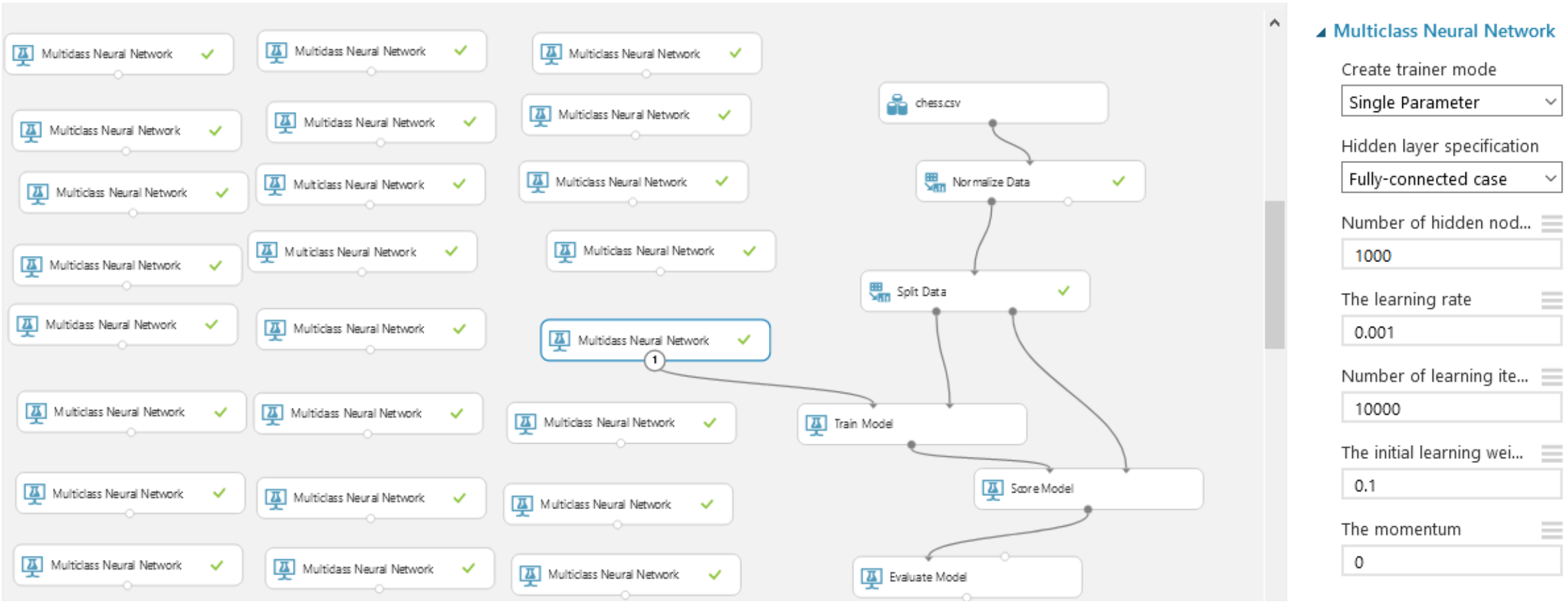

Fig. 18

## Neural Networks with many layers

The Azure Machine Learning tool support defining neural networks with many layers through a simple scripting language called Net#. We used this language to define custom neural networks to increase the accuracy.

| Net# Script | No. of nodes | Learning Rate | No. of iterations | Overall Accuracy | Average Accuracy |
|---|---|---|---|---|---|
| input Data auto; hidden H [200] from Data all; output Out [18] sigmoid from H all; | 200 | 0.1 | 100 | 0.65 | 0.96 |
| input Data auto; hidden H [200] from Data all; hidden H2 [200] from H all; output Out [18] sigmoid from H2 all; | 400 | 0.1 | 100 | 0.72 | 0.96 |
| input Data auto; hidden H [200] from Data all; hidden H2 [200] from H all; hidden H3 [200] from H2 all; output Out [18] sigmoid from H3 all; | 600 | 0.1 | 100 | 0.77 | 0.97 |
| input Data auto; hidden H [200] from Data all; hidden H2 [200] from H all; hidden H3 [200] from H2 all; hidden H4 [200] from H3 all; | 800 | 0.1 | 100 | 0.76 | 0.97 |
| output Out [18] sigmoid from H4 all; | | | | | |
| input Data auto;<br>hidden H [1000] from Data all;<br>hidden H2 [1000] from H all;<br>hidden H3 [1000] from H2 all;<br>output Out [18] sigmoid from H3 all; | 3000 | 0.1 | 100 | 0.85 | 0.98 |
| input Data auto;<br>hidden H [3000] from Data all;<br>hidden H2 [3000] from H all;<br>hidden H3 [3000] from H2 all;<br>output Out [18] sigmoid from H3 all; | 9000 | 0.1 | 100 | 0.85 | 0.98 |
| input Data auto;<br>hidden H [200] from Data all;<br>hidden H2 [200] from H all;<br>hidden H3 [200] from H2 all;<br>output Out [18] sigmoid from H3 all; | 600 | 0.01 | 100 | 0.66 | 0.96 |
| input Data auto;<br>hidden H [1000] from Data all;<br>hidden H2 [1000] from H all;<br>hidden H3 [1000] from H2 all;<br>output Out [18] sigmoid from H3 all; | 3000 | 0.01 | 100 | 0.62 | 0.95 |
| input Data auto;<br>hidden H [200] from Data all;<br>hidden H2 [200] from H all;<br>hidden H3 [200] from H2 all;<br>output Out [18] sigmoid from H3 all; | 600 | 0.1 | 1000 | 0.83 | 0.98 |
| input Data auto;<br>hidden H [200] from Data all;<br>hidden H2 [200] from H all;<br>hidden H3 [200] from H2 all;<br>output Out [18] sigmoid from H3 all; | 600 | 0.1 | 2000 | 0.83 | 0.98 |

Table. 5

We got the highest accuracy using 3000 nodes distributed on three hidden layers where each layer contains 1000 nodes. The activation function is Sigmod and the learning rate is 0.1 while the number of learning iterations is 100. The next screen shot demonstrates the selection of this neural network. The Net# script is written in the properties panel on the right side of the screen.

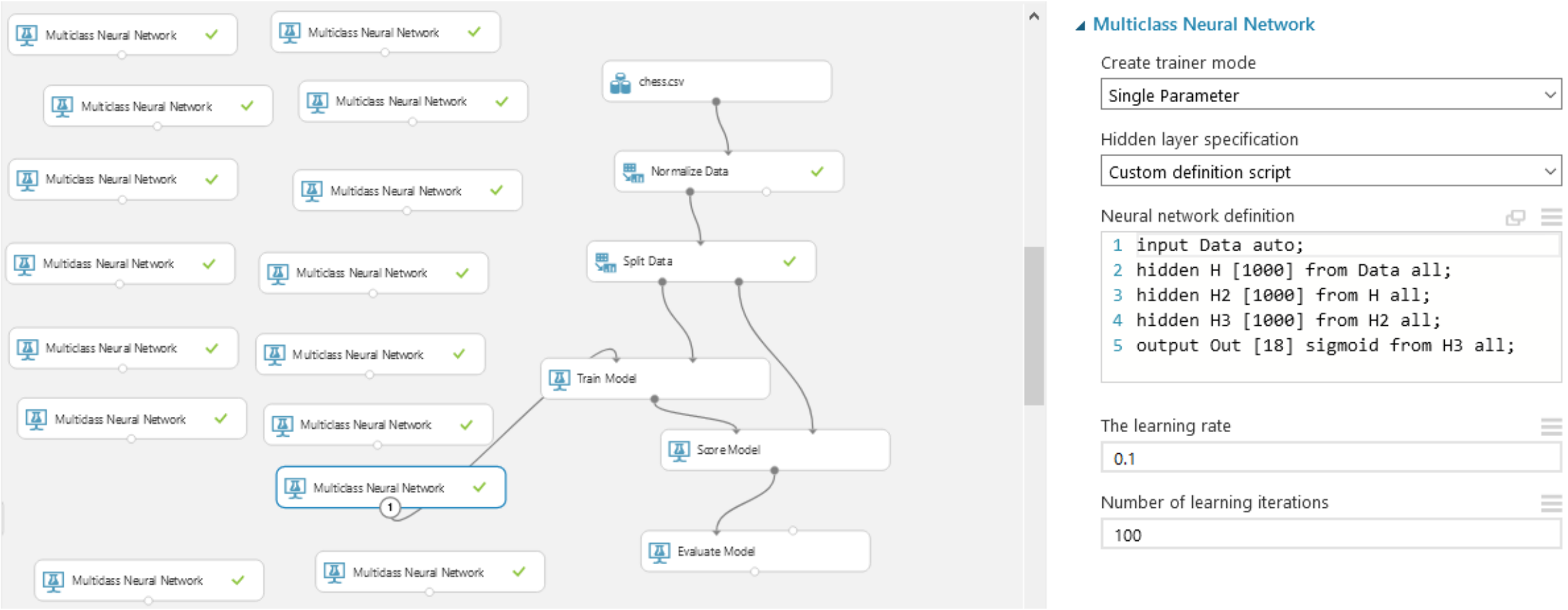

Fig. 19

## The Final Results

The next table presents each model, the overall accuracy, and the average accuracy.

The models are list descending from the highest accuracy to the lowest one.

| Model | Overall Accuracy | Average Accuracy |
|---|---|---|
| Neural Networks | 0.85 | 0.98 |
| Decision Forest | 0.79 | 0.97 |
| Decision Jungle | 0.49 | 0.94 |
| Logistic Regression | 0.32 | 0.92 |

Table. 6

We get the highest accuracy using the Neural Networks model (85%) then the Decision forest model (79%). With respect to the Training time, the Neural Networks consumes larger time (13 hours) while the other models consume (2-3 mins). Based on these results, we encourage using Neural Networks for this type of problems when we care on the best accuracy if the training time is not a problem. In case someone wants to develop something on the fly, like doing a live demonstration without earlier preparation, or doing some quick tests, we recommend using a Decision Forest model for a short training time, and still provide good accuracy. Then switch to Neural Networks if the work is going to be more serious and will be used in production.

**Web Services**

We used the Microsoft Azure Machine Learning to provide a web service for each trained model (Logistic Regression, Decision Jungle, Decision Forest & Neural Networks)

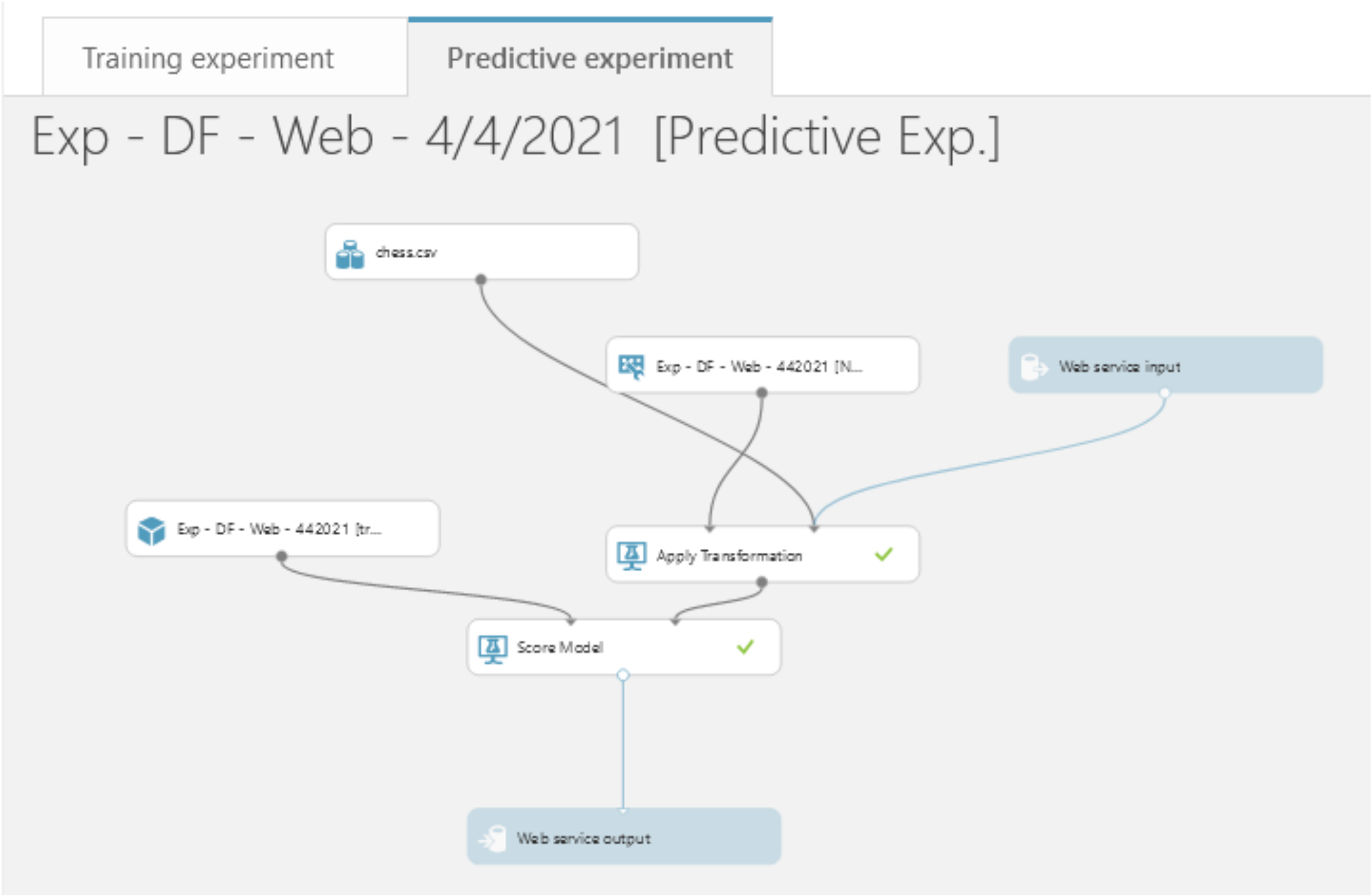

Fig. 20

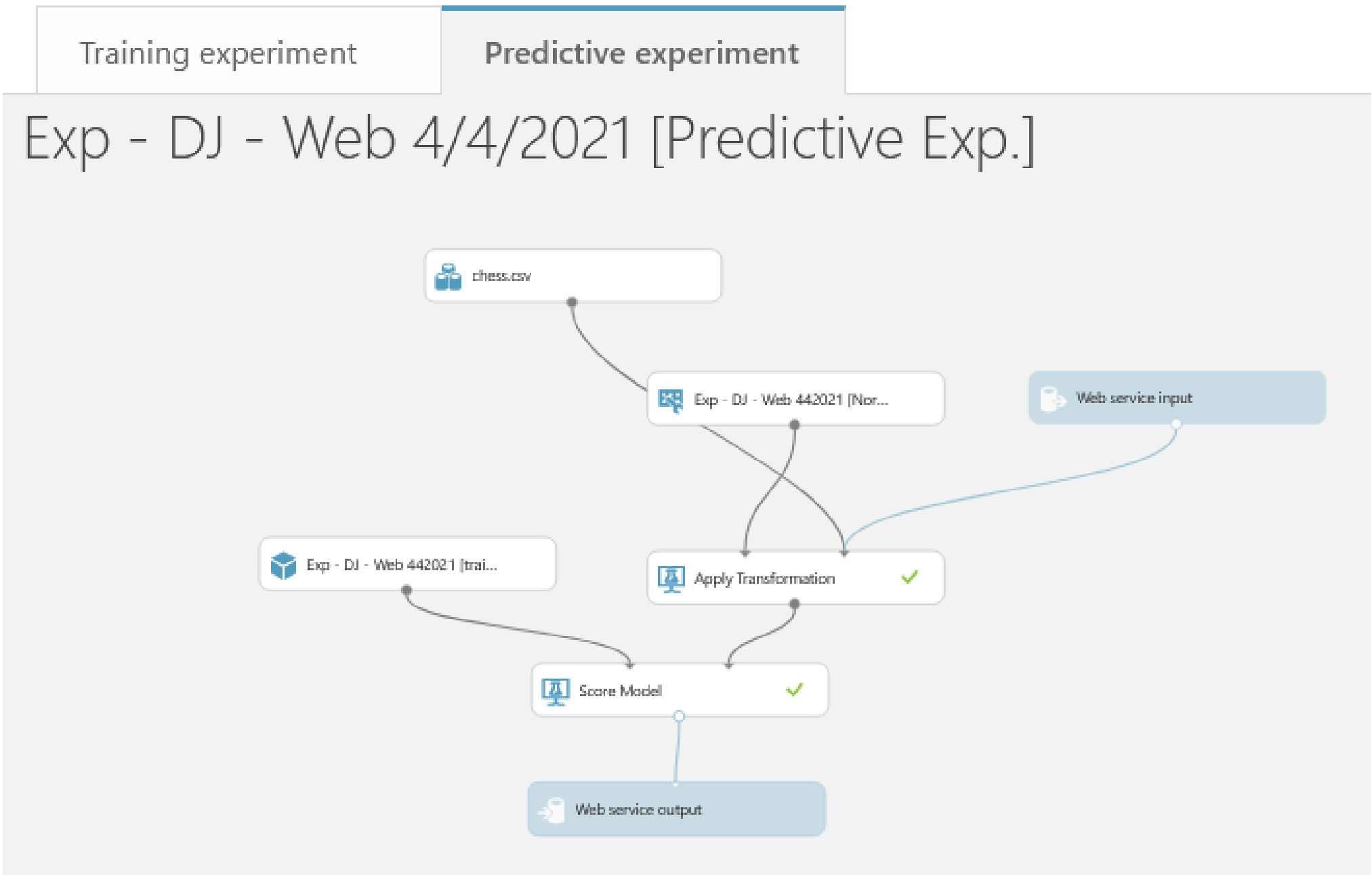

Fig. 21

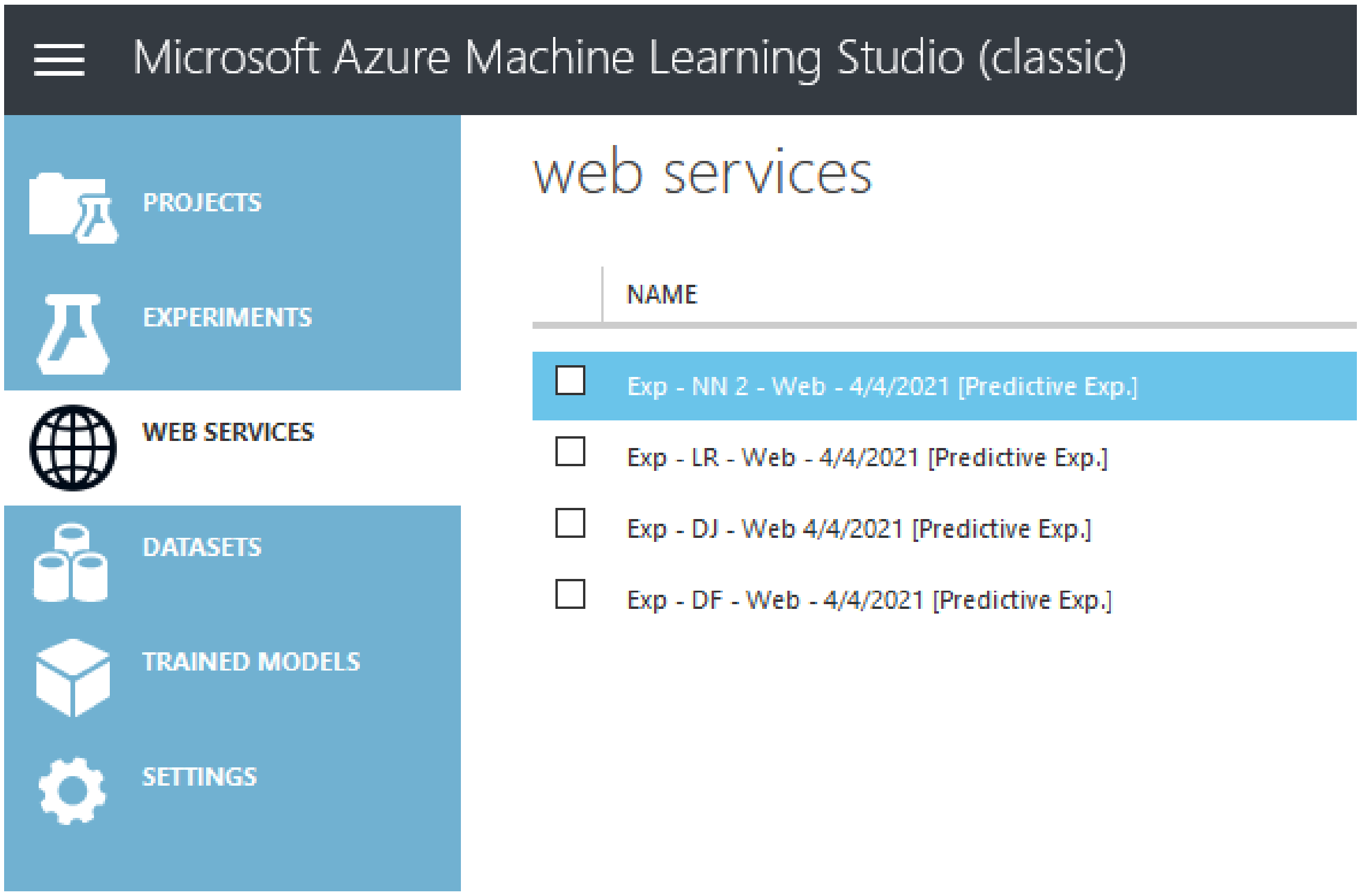

Fig. 22

We can use these web services from the Web Interface provided by Microsoft Azure Machine Learning, or we can consume it using any programming language that support calling web services.

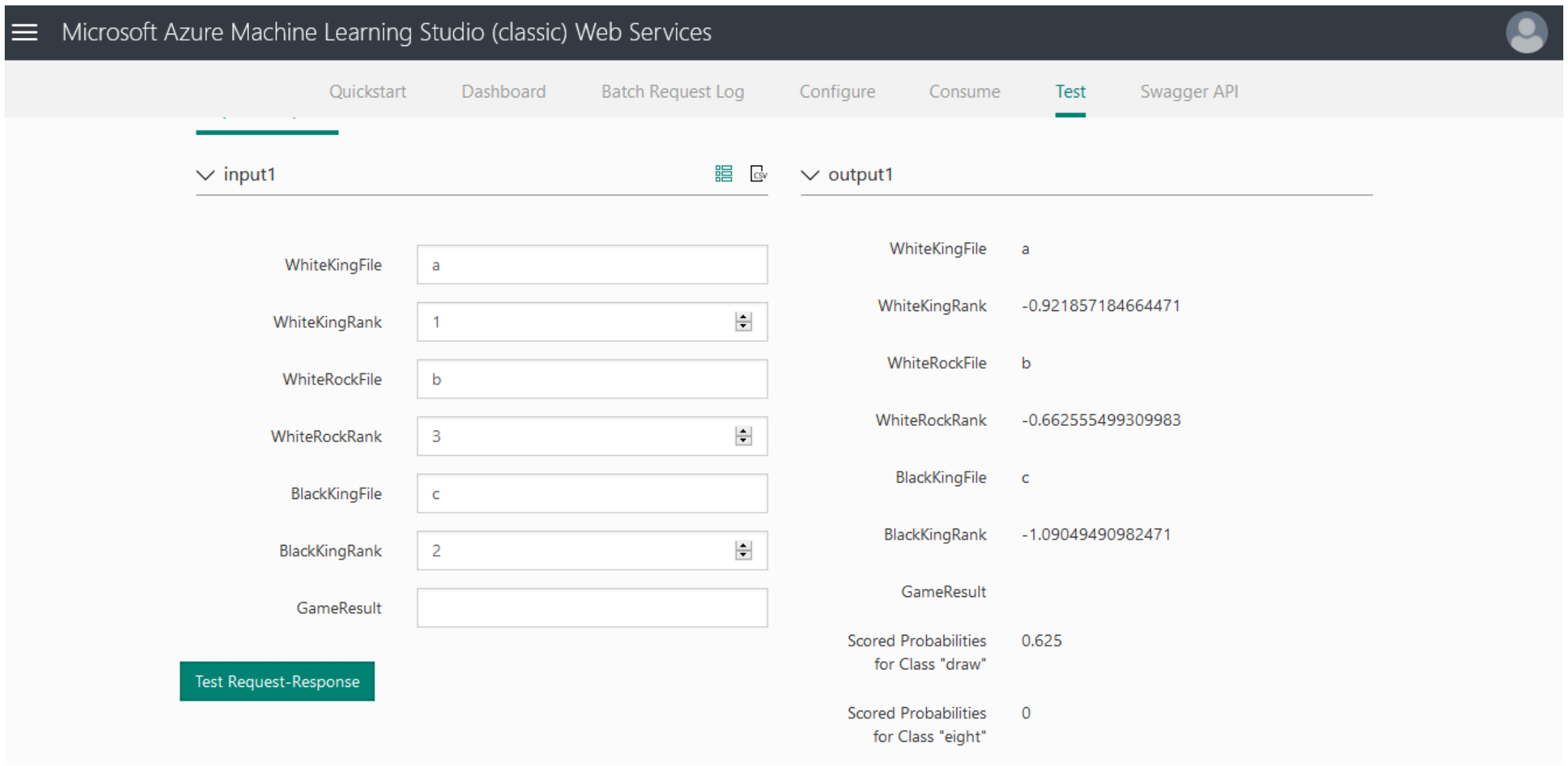

Fig. 23

## Chess Endgame Data Set Application

The application comes with the next features.

- Data Set Visualization
- Data Set Statistics
- Prediction using Logistic Regression, Decision Jungle, Decision Forest, and Neural Networks

The next screen shot demonstrates the Data Set Visualization. We have 28056 samples in the data set, we can select any sample to see the chess board and pieces in the correct position according to the sample. The Visualization is fast a dynamic, once we change the selected sample, the chess board is updated automatically. In the chess board we have the Black King, White King and the White Rook. The Game Result could be Draw, Zero, One, Two, …., Sixteen.

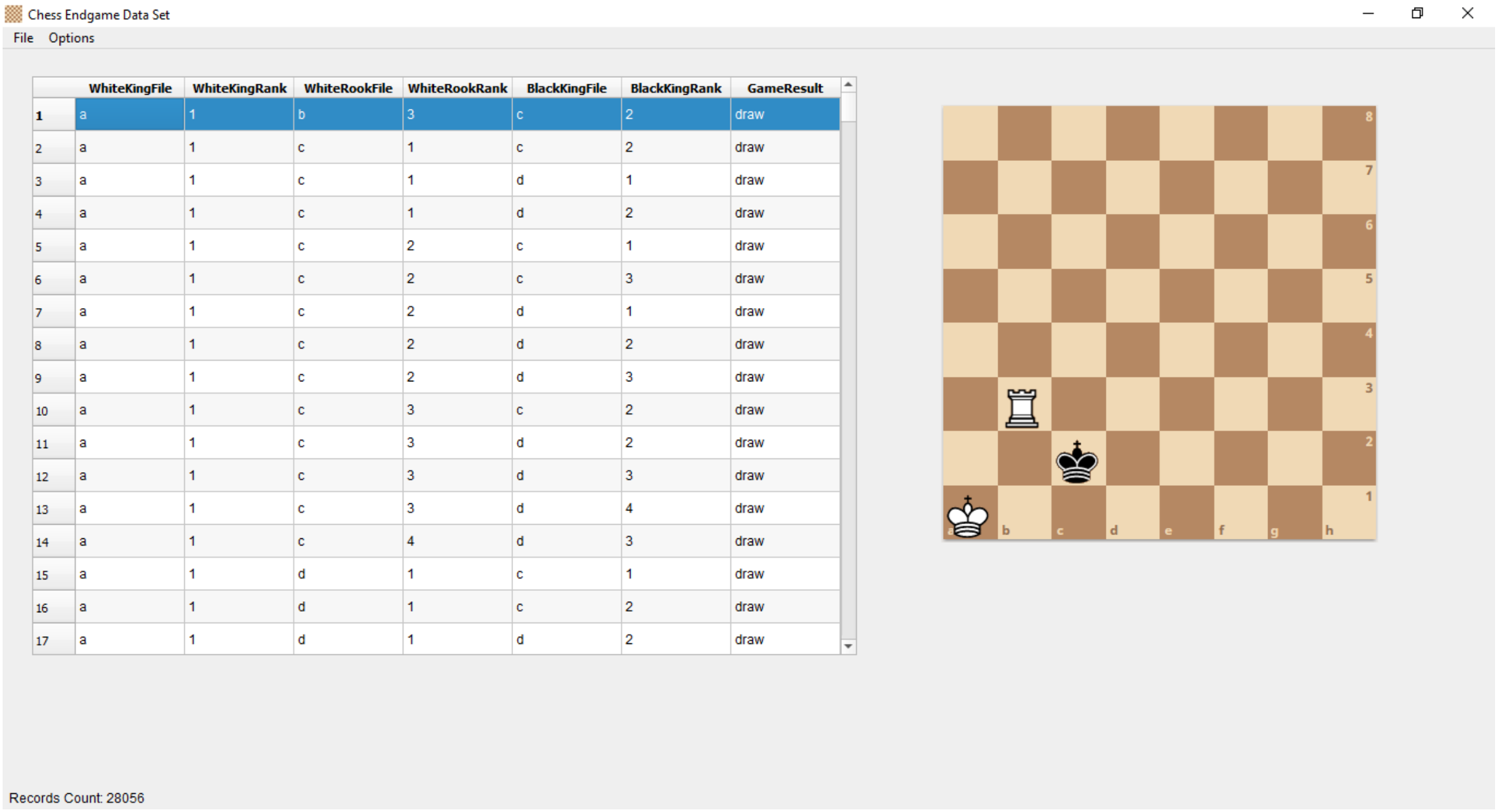

| | WhiteKingFile | WhiteKingRank | WhiteRookFile | WhiteRookRank | BlackKingFile | BlackKingRank | GameResult |
|---|---|---|---|---|---|---|---|
| 1 | a | 1 | b | 3 | c | 2 | draw |
| 2 | a | 1 | c | 1 | c | 2 | draw |
| 3 | a | 1 | c | 1 | d | 1 | draw |
| 4 | a | 1 | c | 1 | d | 2 | draw |
| 5 | a | 1 | c | 2 | c | 1 | draw |
| 6 | a | 1 | c | 2 | c | 3 | draw |
| 7 | a | 1 | c | 2 | d | 1 | draw |
| 8 | a | 1 | c | 2 | d | 2 | draw |
| 9 | a | 1 | c | 2 | d | 3 | draw |
| 10 | a | 1 | c | 3 | c | 2 | draw |
| 11 | a | 1 | c | 3 | d | 2 | draw |
| 12 | a | 1 | c | 3 | d | 3 | draw |
| 13 | a | 1 | c | 3 | d | 4 | draw |
| 14 | a | 1 | c | 4 | d | 3 | draw |
| 15 | a | 1 | d | 1 | c | 1 | draw |
| 16 | a | 1 | d | 1 | c | 2 | draw |
| 17 | a | 1 | d | 1 | d | 2 | draw |

Fig. 24

The next screen shot demonstrates the Data Set statistics window. We have a Table that present the Game Result, Count and Percentage. For example, we have 2796 samples where the Game Result is a Draw which represent 9.97% of the samples.

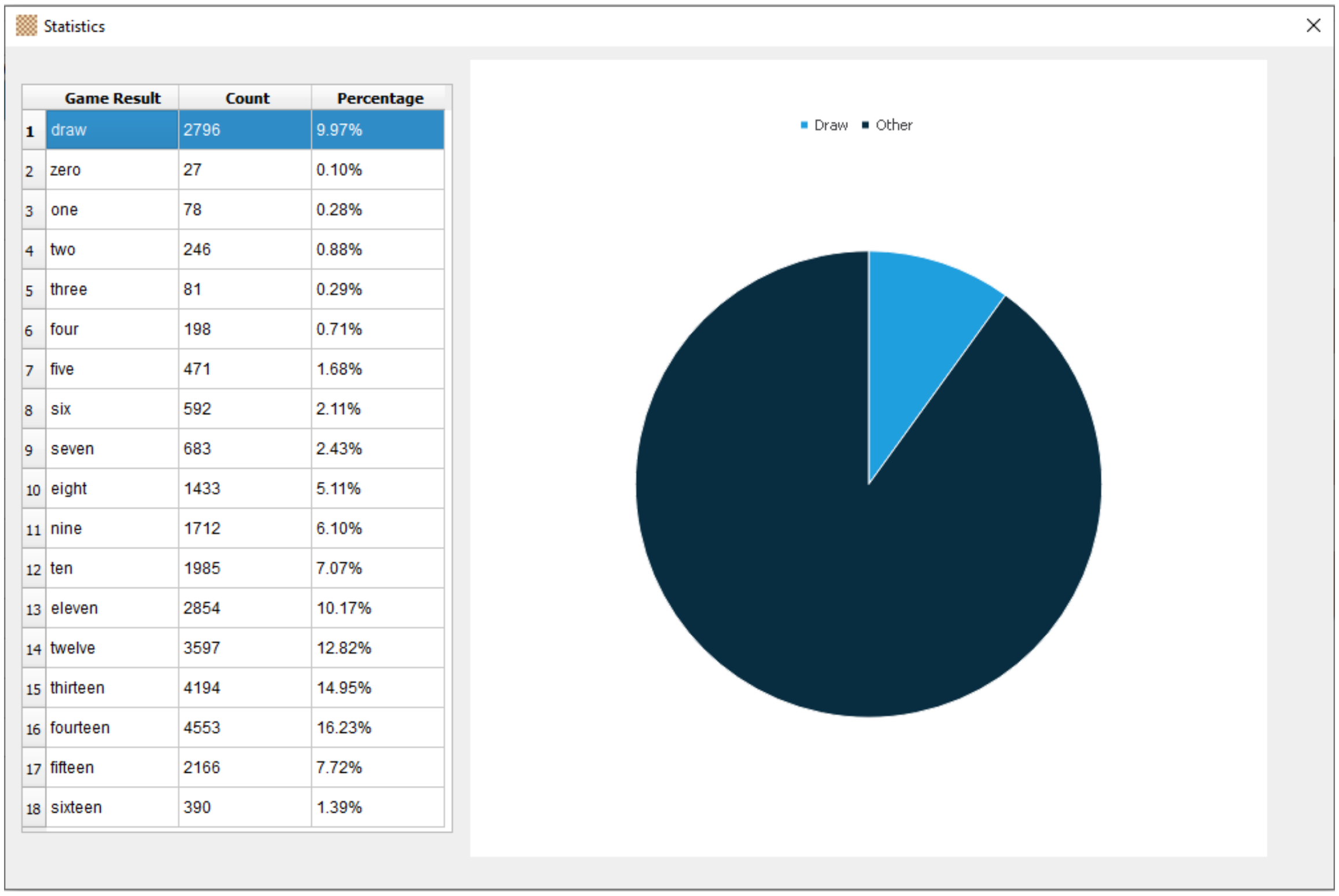

Statistics

| | Game Result | Count | Percentage |
|---|---|---|---|
| 1 | draw | 2796 | 9.97% |
| 2 | zero | 27 | 0.10% |
| 3 | one | 78 | 0.28% |
| 4 | two | 246 | 0.88% |
| 5 | three | 81 | 0.29% |
| 6 | four | 198 | 0.71% |
| 7 | five | 471 | 1.68% |
| 8 | six | 592 | 2.11% |
| 9 | seven | 683 | 2.43% |
| 10 | eight | 1433 | 5.11% |
| 11 | nine | 1712 | 6.10% |
| 12 | ten | 1985 | 7.07% |
| 13 | eleven | 2854 | 10.17% |
| 14 | twelve | 3597 | 12.82% |
| 15 | thirteen | 4194 | 14.95% |
| 16 | fourteen | 4553 | 16.23% |
| 17 | fifteen | 2166 | 7.72% |
| 18 | sixteen | 390 | 1.39% |

Fig. 25

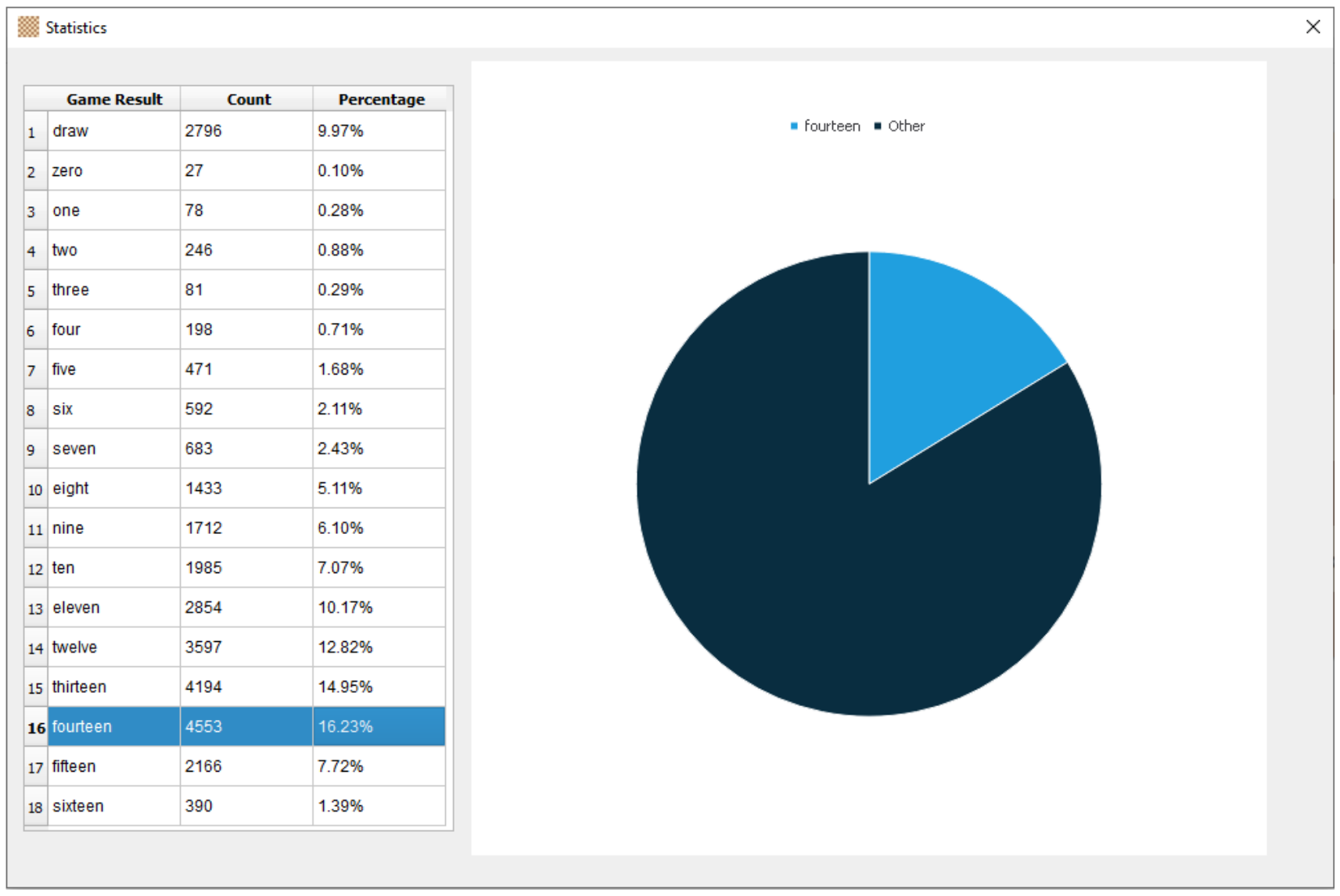

Statistics

| | Game Result | Count | Percentage |
|---|---|---|---|
| 1 | draw | 2796 | 9.97% |
| 2 | zero | 27 | 0.10% |
| 3 | one | 78 | 0.28% |
| 4 | two | 246 | 0.88% |
| 5 | three | 81 | 0.29% |
| 6 | four | 198 | 0.71% |
| 7 | five | 471 | 1.68% |
| 8 | six | 592 | 2.11% |
| 9 | seven | 683 | 2.43% |
| 10 | eight | 1433 | 5.11% |
| 11 | nine | 1712 | 6.10% |
| 12 | ten | 1985 | 7.07% |
| 13 | eleven | 2854 | 10.17% |
| 14 | twelve | 3597 | 12.82% |
| 15 | thirteen | 4194 | 14.95% |
| 16 | fourteen | 4553 | 16.23% |
| 17 | fifteen | 2166 | 7.72% |
| 18 | sixteen | 390 | 1.39% |

Fig. 26

The next screen shots for the Prediction window. In this window we can set the board parameters (White King file & rank, Blank King file & Rank and the White Rook file & rank). This sample could exist in the Data Set or not. Then we select the Algorithm (Multiclass Decision Forest, Multiclass Decision Jungle, Neural Network or Logistic Regression).

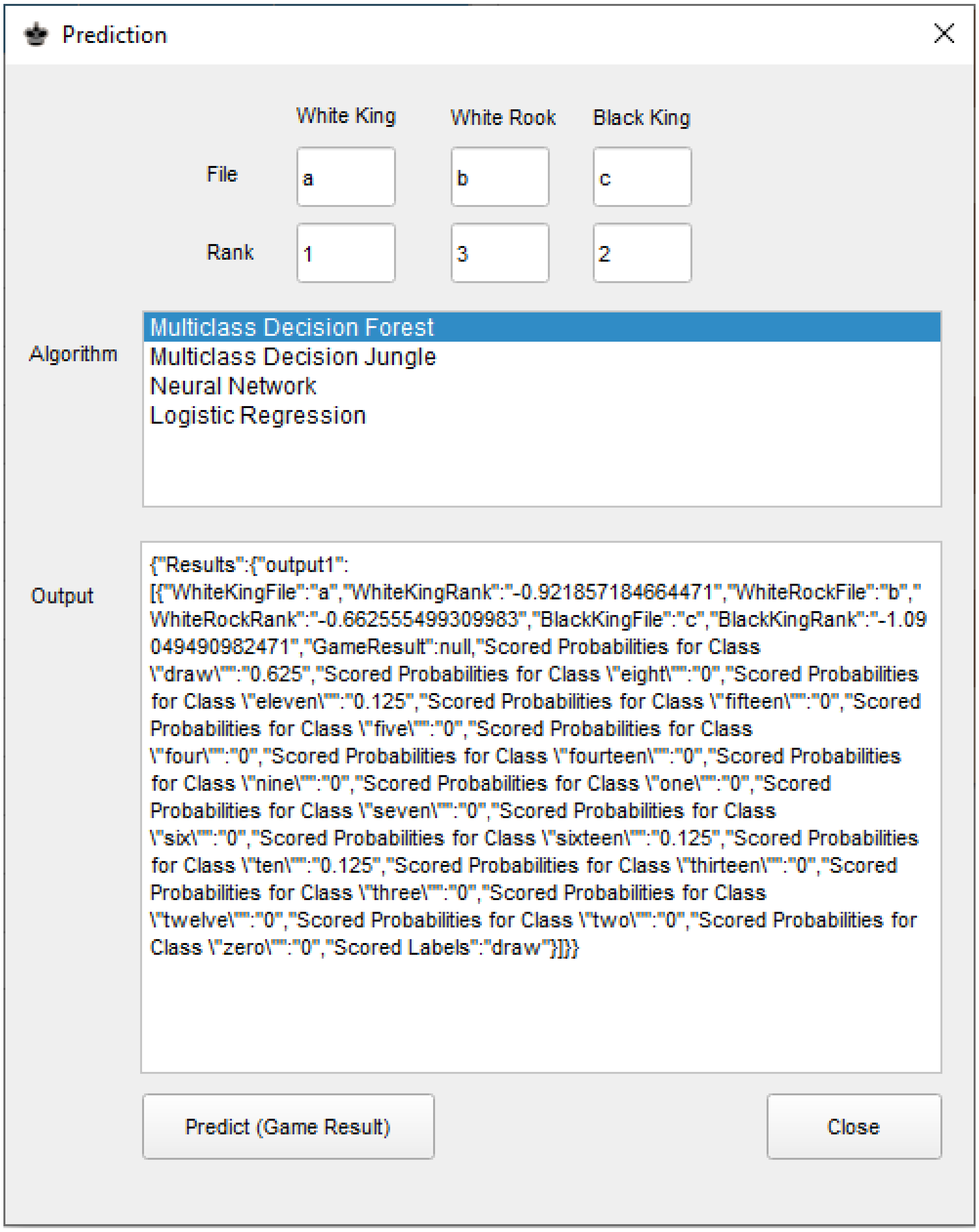

Fig. 27

## Future Work

In the future we will extend our experiments, for example we will try more neural networks with different scripts that set the layers count, nodes in each layer, different connects, and different activation functions. Also, we plan to try different models like the Support Vector Machine (SVM). We plan also to try ensemble learning and use many different models together in the prediction process to get higher accuracy. An improvement that we plan to do too is developing a tool that provide a simple GUI to design the Neural Networks architecture and generate the Net# source code for the Microsoft Azure Machine Learning tool. We plan also to try the same algorithms with different Data Sets and see the change in the accuracy and analysis the results.

## Conclusion

In this paper we presented a Machine Learning solution to predict the result of the Chess Endgame (King-Rook vs King) problem using different algorithms like Neural Networks, Logistic Regression, Decision Jungle and Decision Forest. We did many experiments to calculate the Accuracy of each model and determine which one provides the highest accuracy. Out results demonstrate that using Neural Networks we can reach 85% accuracy after 13 hours of training time using Microsoft Azure Machine Learning tool. We also developed an application using the Ring programming language that visualize the data set and predict the game result using different algorithms.

## References

[1] Munshi, Jamal. "Pairwise comparison of chess opening variations." Available at SSRN 2472783 (2014).

[2] Znosko-Borovsky, Eugene, and Evgeniĭ Aleksandrovich Znosko-Borovskiĭ. The middle game in chess. Courier Corporation, 1980

[3] Turing, Alan M. "Chess." Computer Chess Compendium. Springer, New York, NY, 1988. 14-17.

[4] Basic Chess Endgames, Ruben Fine & revised by Pal Benko, 2003

[5] A Pocket Guide to Endgames, David Hooper, 1970

[6] Hauptman, Ami, and Moshe Sipper. "GP-endchess: Using genetic programming to evolve chess endgame players." European Conference on Genetic Programming. Springer, Berlin, Heidelberg, 2005.

[7] Shapiro, Alen, and Tim Niblett. "Automatic induction of classification rules for a chess endgame." Advances in computer chess. Pergamon, 1982. 73-92.

[8] De Bruin, Anique BH, Remy MJP Rikers, and Henk G. Schmidt. "Monitoring accuracy and self-regulation when learning to play a chess endgame." Applied Cognitive Psychology 19.2 (2005): 167-181.

[9] Aly, Mohamed. "Survey on multiclass classification methods." Neural Netw 19 (2005): 1-9.

[10] Tewari, Ambuj, and Peter L. Bartlett. "On the Consistency of Multiclass Classification Methods." Journal of Machine Learning Research 8.5 (2007).

[11]https://archive.ics.uci.edu/ml/datasets/Chess+%28King-Rook+vs.+King%29

[12] M. Bain. "Learning optimal chess strategies", ILP 92: ICOT TM-1182, S. Muggleton, Institute for New Generation Computer Technology, Tokyo, Japan.

[13] M. Bain. Learning Logical Exceptions in Chess. PhD dissertation. University of Strathclyde. 1994.

[14] Muggleton, Stephen, ed. Inductive logic programming. No. 38. Morgan Kaufmann, 1992

[15] Lavrac, Nada, and Saso Dzeroski. "Inductive Logic Programming." WLP. 1994.

[16] Muggleton, Stephen, and Luc De Raedt. "Inductive logic programming: Theory and methods." The Journal of Logic Programming 19 (1994): 629-679.

[17] Lassabe, Nicolas, et al. "Genetically programmed strategies for chess endgame." Proceedings of the 8th annual conference on genetic and evolutionary computation. 2006.

[18] Wayne Iba, Searching for Better Performance on the King-Rook-King Chess Endgame Problem, In Proceedings of the Twenty-fifth International FLAIRS Conference (2012).

[19] Gasso, Gilles. "Logistic regression." (2019).

[20] Matsui, Hidetoshi. "Variable and boundary selection for functional data via multiclass logistic regression modeling." Computational Statistics & Data Analysis 78 (2014): 176-185.

[21] Gunarathne, W. H. S. D., K. D. M. Perera, and K. A. D. C. P. Kahandawaarachchi. "Performance evaluation on machine learning classification techniques for disease classification and forecasting through data analytics for chronic kidney disease (CKD)." 2017 IEEE 17th international conference on bioinformatics and bioengineering (BIBE). IEEE, 2017.

[22] Rahnama, Alireza, Guilherme Zepon, and Seetharaman Sridhar. "Machine learning based prediction of metal hydrides for hydrogen storage, part II: Prediction of material class." International Journal of Hydrogen Energy 44.14 (2019): 7345-7353.

[23] Hong, Huixiao, et al. "Multiclass Decision Forest—a novel pattern recognition method for multiclass classification in microarray data analysis." DNA and cell biology 23.10 (2004): 685-694.

[24] Hong, Huixiao, et al. "Development of decision forest models for prediction of drug-induced liver injury in humans using a large set of FDA-approved drugs." Scientific reports 7.1 (2017): 1-15.

[25] Rajagopal, Smitha, Katiganere Siddaramappa Hareesha, and Poornima Panduranga Kundapur. "Performance analysis of binary and multiclass models using azure machine learning." International Journal of Electrical & Computer Engineering (2088-8708) 10.1 (2020).

[26] Kröse, Ben, et al. "An introduction to neural networks." (1993).

[27] Gurney, Kevin. An introduction to neural networks. CRC press, 1997.

[28] Yegnanarayana, Bayya. Artificial neural networks. PHI Learning Pvt. Ltd., 2009.

[29] Bishop, Chris M. "Neural networks and their applications." Review of scientific instruments 65.6 (1994): 1803-1832.

[30] Hassoun, Mohamad H. Fundamentals of artificial neural networks. MIT press, 1995.

[31] Hecht-Nielsen, Robert. "Theory of the backpropagation neural network." Neural networks for perception. Academic Press, 1992. 65-93.

[32] Widrow, Bernard, and Michael A. Lehr. "30 years of adaptive neural networks: perceptron, madaline, and backpropagation." Proceedings of the IEEE 78.9 (1990): 1415-1442.

[33] Burnett, Margaret M., and David W. McIntyre. "Visual programming." COMPUTER-LOS ALAMITOS- 28 (1995): 14-14.

[34] Myers, Brad A. "Taxonomies of visual programming and program visualization." Journal of Visual Languages & Computing 1.1 (1990): 97-123.

[35] Barga, Roger, et al. Predictive analytics with Microsoft Azure machine learning. Berkely, CA: Apress, 2015.

[36] Mund, Sumit. Microsoft azure machine learning. Packt Publishing Ltd, 2015.

[37] Barnes, Jeff. Microsoft Azure Essentials Azure Machine Learning. Microsoft Press, 2015.

[38] Ayouni, Mansour. Beginning Ring Programming. Apress, 2020.

[39] Fayed, Mahmoud S., et al. "PWCT: visual language for IoT and cloud computing applications and systems." Proceedings of the Second International Conference on Internet of things, Data and Cloud Computing. 2017.

[40] Fayed, Mahmoud S., et al. "PWCT: a novel general-purpose visual programming language in support of pervasive application development." CCF Transactions on Pervasive Computing and Interaction 2.3 (2020): 164-177.

[41] Fayed, M.S., 2017. General-Purpose Visual Language and Information System with Case-Studies in Developing Business Applications. arXiv preprint arXiv:1712.10281.

[42] Imran, M., Alnuem, M.A., Fayed, M.S. and Alamri, A., 2013. Localized algorithm for segregation of critical/non-critical nodes in mobile ad hoc and sensor networks. Procedia Computer Science, 19, pp.1167-1172

[43] Alnuem, M., Zafar, N.A., Imran, M., Ullah, S. and Fayed, M., 2014. Formal specification and validation of a localized algorithm for segregation of critical/noncritical nodes in MAHSNs. International Journal of Distributed Sensor Networks, 10(6), p.140973

*[44]* Ring Team (4 December 2017). *"Ring and other languages"*. ring-lang.net. *ring-lang*. *Archived* from the original on 25 December 2018. *Retrieved 4 December 2017*.